\documentclass[10pt,twocolumn,letterpaper]{article}

\usepackage{iccv}
\usepackage{times}
\usepackage{epsfig}
\usepackage{graphicx}
\usepackage{amsmath}
\usepackage{amssymb}
\usepackage{graphics}
\usepackage{subfigure}
\usepackage{amssymb}
\usepackage{amsthm}
\usepackage{amsmath}
\usepackage{algorithm}
\usepackage{algorithmic}

\usepackage{booktabs}
\usepackage[table]{xcolor}

\definecolor{yelloworange}{RGB}{255, 153, 0}
\definecolor{ultramarineblue}{RGB}{65, 102, 245}

\usepackage[pagebackref=true,breaklinks=true,letterpaper=true,colorlinks,bookmarks=false]{hyperref}

\iccvfinalcopy 

\def\iccvPaperID{****} 

\begin{document}

\title{Fully Convolutional Attention Networks for Fine-Grained Recognition}

\author{Xiao Liu, Tian Xia, Jiang Wang, Yi Yang, Feng Zhou and Yuanqing Lin\\
Baidu Research\\
{\tt\small \{liuxiao12,xiatian,wangjiang03,yangyi05, zhoufeng09, linyuanqing\}@baidu.com}
}

\maketitle

\begin{abstract}
Fine-grained recognition is challenging due to its subtle local inter-class differences versus large intra-class variations such as poses.
A key to address this problem is to localize discriminative parts to extract pose-invariant features.
However, ground-truth part annotations can be expensive to acquire.
Moreover, it is hard to define parts for many fine-grained classes.
This work introduces \textbf{Fully Convolutional Attention Networks (FCANs)}, a reinforcement learning framework to optimally glimpse local discriminative regions adaptive to different fine-grained domains.
Compared to previous methods, our approach enjoys three advantages:
1) the weakly-supervised reinforcement learning procedure requires no expensive part annotations;
2) the fully-convolutional architecture speeds up both training and testing;
3) the greedy reward strategy accelerates the convergence of the learning.
We demonstrate the effectiveness of our method with extensive experiments on four challenging fine-grained benchmark datasets, including CUB-200-2011, Stanford Dogs, Stanford Cars and Food-101.
\end{abstract}

\section{Introduction}
\begin{figure}
\begin{center}
\includegraphics[width=.5\textwidth]{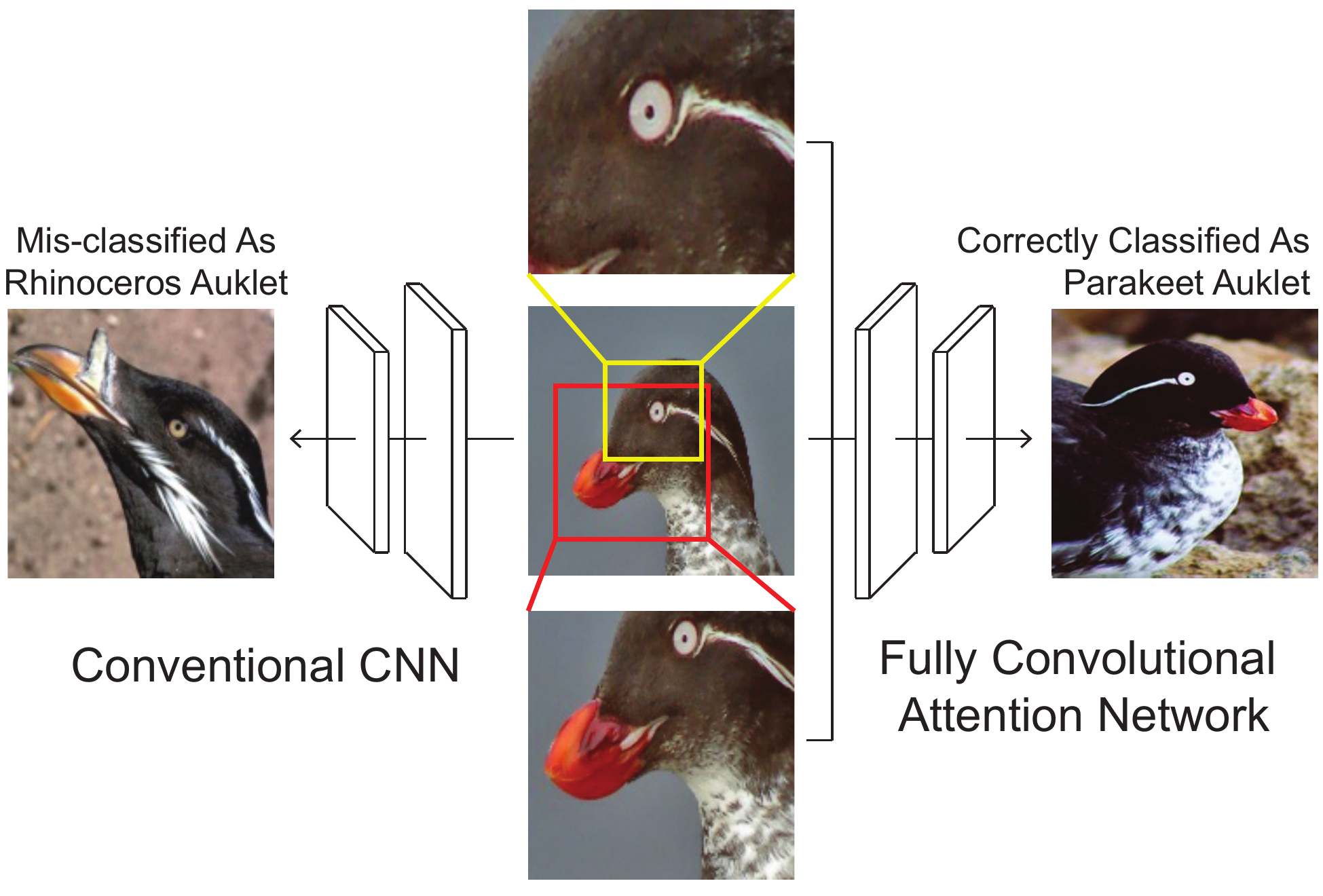}
\end{center}
\caption{Conventional CNN approach (left) finds difficulty in differentiating similar fine-grained categories with subtle local variations (eg., Rhinoceros Auklet against Parakeet Auklet).
In contrast, our proposed fully convolutional attention networks (right) is able to automatically and efficiently localize parts (eg., bird's eye and beak) given only weakly supervised fine-grained class labels.
}
\label{fig:teaser}
\end{figure}

Fine-grained recognition refers to the task of distinguishing sub-ordinate categories, such as bird species~\cite{wah2011caltech}, dog breeds~\cite{khosla2011novel}, car models~\cite{krause20133d}, flower categories~\cite{nilsback2008automated}, food dishes~\cite{bossard2014food}, etc.
With the great potential in rivaling human experts, it has shown tremendous applications in real world ranging from e-commerce~\cite{bell2015learning, hadi2015buy} to education~\cite{kumar2012leafsnap, berg2014birdsnap}.
Although great success has been achieved for basic-level recognition in the last few years~\cite{krizhevsky2012imagenet, simonyan2014very, szegedy2015going, he2016deep}, fine-grained recognition still faces two challenges.
First, it is more difficult and time-consuming to gather a large amount of labeled fine-grained data because it calls for experts with specialized domain knowledge.
In addition, the difference between fine-gained classes is very subtle.
The most discriminative features are often not based on the global shape or appearance variation but contained in the mis-alignment of local parts or patterns.
For instance, as shown in Fig.~\ref{fig:teaser}, the eye texture and beak shape are crucial to differentiate between Parakeet Auklet and Rhinoceros Auklet.

\setlength{\tabcolsep}{0.5pt}
\begin{figure}[t]
\begin{center}
\begin{tabular}{ccc}
\includegraphics[height=0.25\linewidth,width=0.25\linewidth]{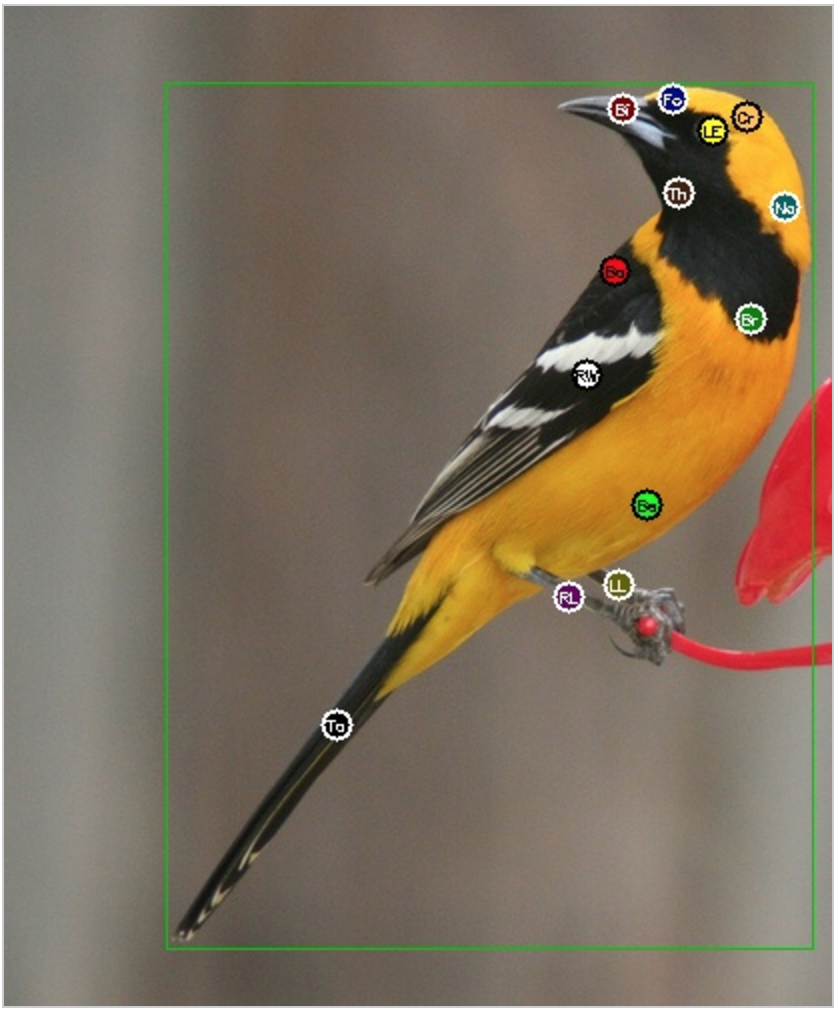} &
\includegraphics[height=0.25\linewidth,width=0.4\linewidth]{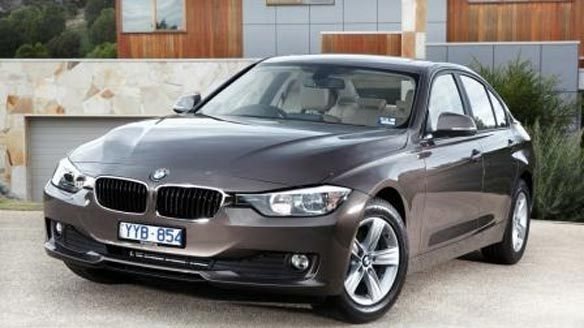} &
\includegraphics[height=0.25\linewidth,width=0.3\linewidth]{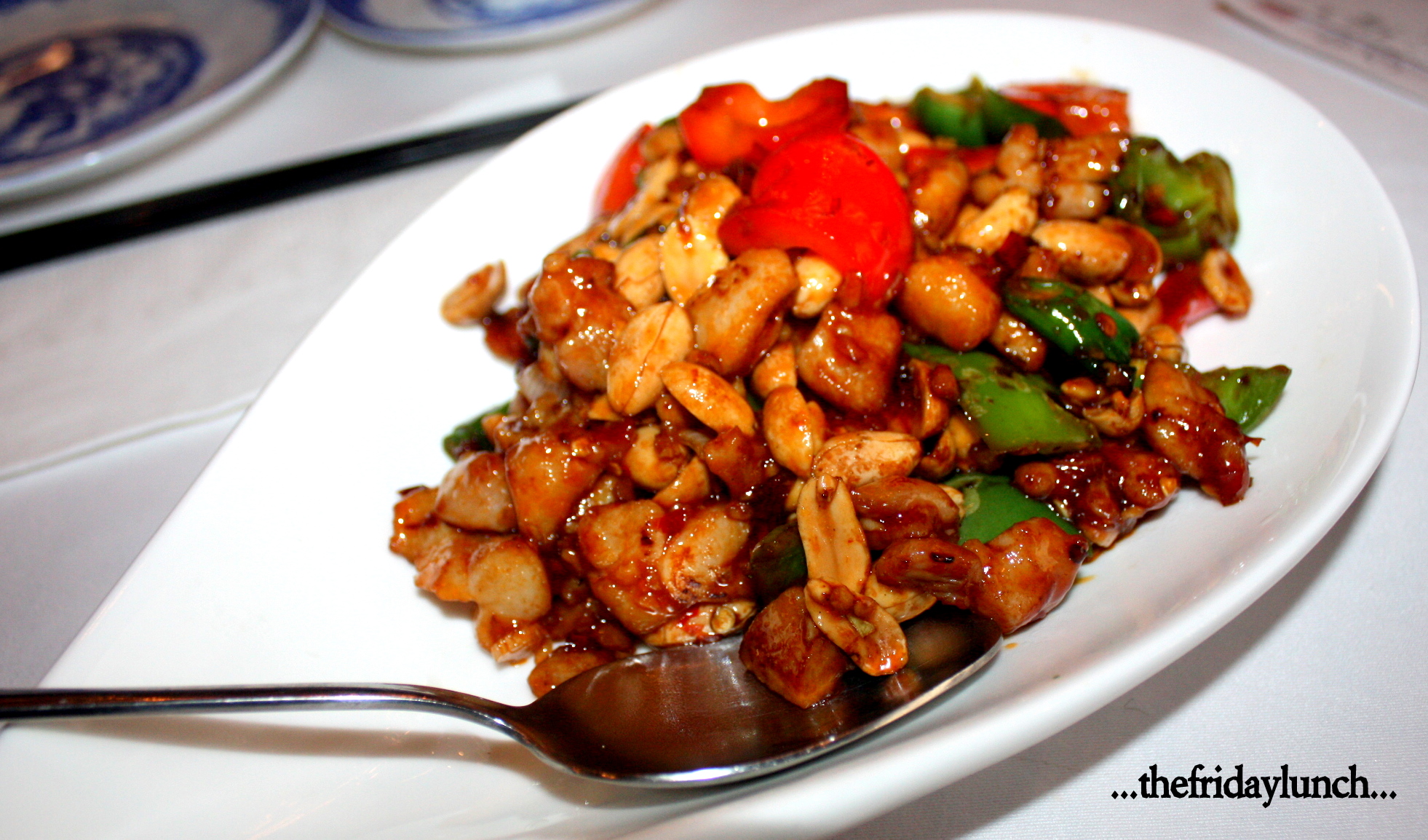} \\
\vspace{2pt}
Hooded Oriole & BMW & Kungpao Chicken \\
\includegraphics[height=0.25\linewidth,width=0.25\linewidth]{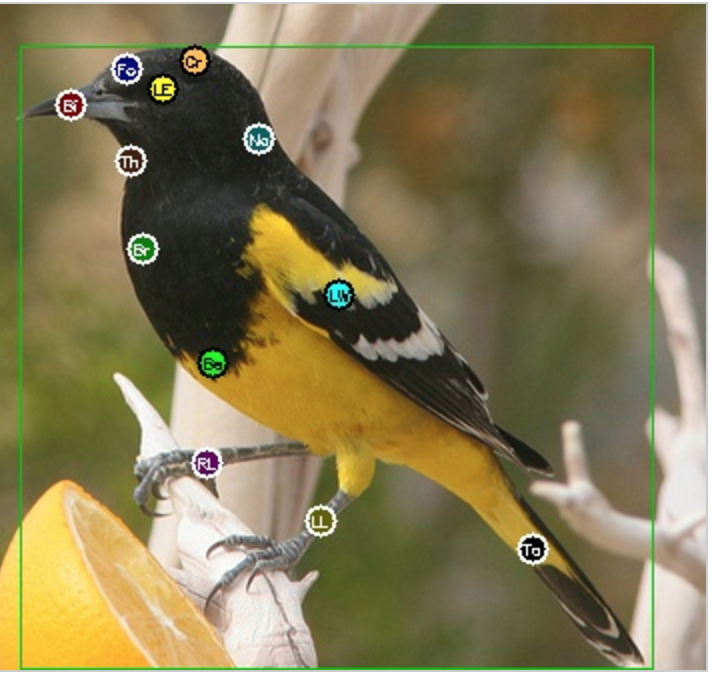} &
\includegraphics[height=0.25\linewidth,width=0.4\linewidth]{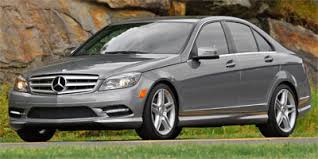} &
\includegraphics[height=0.25\linewidth,width=0.3\linewidth]{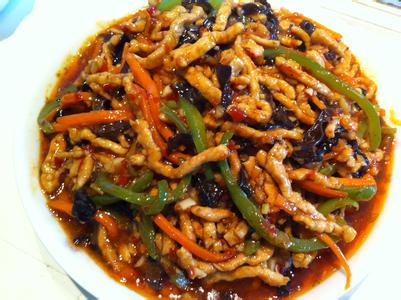} \\
Scott Oriole & Mercedes-Benz & Yuxiang Rousi \\
(a) Bird & (b) Car & (c) Food
\end{tabular}
\end{center}
\caption{Fine-grained recognition often involves part localization, i.e. (a) localizing head and breast to distinguish birds, and (b) localizing brand to classify car makes.
It is relatively easy to define parts for structured objects like birds and cars.
However, it is hard to define rigid parts for unstructured classes, such as (c) food.
This may be solved by attention models.
}
\label{fig:teaser2}
\end{figure}

To that end, the main body of previous research has focused on devising more discriminative features by detecting and aligning object parts.
Nevertheless, most conventional methods~\cite{liu2012dog, branson2014bird} utilize manually defined parts to localize the regions, such as ``the head of a bird'', for fine-grained recognition.
Relying on manually defined parts has several drawbacks:
1) The precise part annotations are usually expensive to acquire.
2) The strongly supervised part-based model might fail if some parts are occluded.
3) For some fine-grained categories, it is very difficult to manually define parts for them.
For example, it is very difficult to define parts for food recognition, as suggested in Fig.~\ref{fig:teaser2}.
4) Most importantly, there is no clue that manually defined parts are optimal for all fine-grained recognition tasks.

To overcome these problems, we propose a visual attention framework called {\em Fully Convolutional Attention Networks} (FCANs) for fine-grained recognition without part annotation.
Given only image label, our framework utilizes reinforcement learning to simultaneously localize object parts and classify the object within the scene.
Intuitively, the framework simulates human visual system that proceeds object recognition via a series of {\em glimpse} on object parts.
At each glimpse, it strives to find the most discriminative location that can differentiate object's category given the previous observations.
Similar to previous visual attention models~\cite{mnih2014recurrent, sermanet2014attention}, we employ the REINFORCE algorithm during training~\cite{williams1992simple}, where the action is the location of each glimpse, the state is the image and the locations of the previous glimpses, and the reward measures the classification correctness.
The whole framework can be trained only by an image classification loss, thus requiring no manual part annotations.
The visual attention approach is demonstrated to perform well on fine-grained recognition without requiring manually labeled object parts~\cite{sermanet2014attention}.

Compared to the previous reinforcement learning-based visual attention frameworks~\cite{mnih2014recurrent, sermanet2014attention}, the FCANs enjoy better computational efficiency as well as higher classification accuracy in fine-grained recognition.
More concretely, our proposed framework improves the attention models in three ways:
\begin{itemize}
\item{\bf Computational Efficiency:} The previous frameworks run a convolutional neural network individually on each image crop, which is computationally expensive during both training and testing.
In contrast, our method re-uses the same feature maps (computed by a fully convolutional neural network~\cite{szegedy2015going, simonyan2014very}) during each glimpse in a way similar to Fast-RCNN~\cite{girshick2015fast}.
This makes training and prediction computationally more efficient because of the fully convolutional neural network architecture and feature sharing technique.
\item{\bf Multiple Part Localization:} During testing, our model is able to simultaneously locate multiple parts of adaptive sizes, while the previous frameworks~\cite{mnih2014recurrent, sermanet2014attention} generally only locate one part at each iteration.
\item{\bf Faster Training Convergence:} Instead of assigning a delayed reward at the end of attention iterations as previous methods~\cite{mnih2014recurrent, sermanet2014attention}, we apply a new greedy reward strategy at every step of attention, which is crucial to both the convergence speed of training and the accuracy of prediction.
\end{itemize}


As a result, our proposed approach improves the recognition accuracy over previous reinforcement learning based methods~\cite{mnih2014recurrent, sermanet2014attention} while being computationally more efficient.
Our method also achieve competitive results against other state-of-the-art methods on multiple fine-grained datasets.


\section{Related Work}
Fine-grained recognition has been extensively studied in recent years~\cite{bossard2014food, berg2014birdsnap, cui2016fine, huang2016part, krause2015fine, krause20133d, khosla2011novel, liu2012dog, nilsback2008automated}.
We review the three most relevant directions in this section.

\subsection{Representation Learning}
Since the seminal work of AlexNet~\cite{krizhevsky2012imagenet}, we are witnessing a fast-pacing transition from hand-crafted feature to end-to-end convolutional neural networks in representation learning~\cite{simonyan2014very, szegedy2015going, he2016deep}.
Most of the current state-of-the-art fine-grained recognition algorithms are also based on deep CNN representation to distinguish the subtle difference~\cite{gao2016compact, kong2016low, lin2015bilinear}.
Branson \etal~\cite{branson2014bird} claim that integrating lower-level layer and higher-level layer features learns more discriminative representation for fine-grained recognition.
Lin \etal~\cite{lin2015bilinear} propose a bilinear architecture to model local pairwise feature interactions for fine-grained recognition, where convolutional features from two models are combined in a translation invariant manner.
Qian \etal~\cite{qian2015fine} propose a multi-stage metric learning framework to learn a distance metric that pulls data points of the same class close and pushes data points from different classes far apart.
Wang \etal~\cite{wang2014object} combine saliency-aware object detection approach and object-centric sampling scheme to extract more robust and discriminative features for large-scale fine-grained car classification.
In parallel to these efforts, our method combines representation learning with part detection in a unified framework.

\subsection{Part Models}
Since 70's, early cognitive research study~\cite{rosch1976basic} has shown that subordinate-level recognition is based on comparing the appearance details of object parts.
Drawing inspiration from this fact, various pose normalization methods~\cite{farrell2011birdlets, zhang2013deformable, zhang2014panda, liu2012dog, zhang2014part} have been proposed to focus on the important regions.
However, these methods are strongly supervised ones, heavily relying on manually pre-defined parts modeled by Poselet~\cite{bourdev2009poselets} or DPM~\cite{felzenszwalb2010object}.
Due to this limit, most of recent efforts were spent on how to automatically discover critical parts in a weaker setting.
For instance, Berg \etal~\cite{berg2013poof} use data mining techniques to learn a set of intermediate features that can differentiate two classes based on the appearance of a particular part.
Yang \etal~\cite{yang2012unsupervised} propose a template model to discover the common geometric patterns of object parts and the co-occurrence statistics of the patterns.
Similarly, Gavves \etal~\cite{gavves2013fine} and Chai \etal~\cite{chai2011bicos} segment images and align the image segments in an unsupervised fashion.
The aligned image segments are utilized for feature extraction separately.
Recently, Simon and Rodner~\cite{simon2015neural} propose neural activation constellations, an approach that is able to learn part models in an unsupervised manner.
Compared to our method, however, these methods require tedious and ad-hoc tuning of individual components.



\subsection{Attention Models}
One of the main drawbacks of part-based models is the need for a strong motivation in part definition (either by hand or by data-mining method), which may lack for many non-structured objects such as food dishes~\cite{krause2016unreasonable}.
On the other hand, several works introduce attention-based models for task-driven object/part localization.
For instance, Mnih \etal~\cite{mnih2014recurrent} present a recurrent neural network model for object detection by adaptively selecting a sequence of attention regions and extract appearance representations in these regions.
Since this model is non-differentiable, it is trained with reinforcement learning technique to learn task-specific policies.
Ba \etal~\cite{ba2014multiple} extend~\cite{mnih2014recurrent} and successfully achieve good results on a more challenging multi-digit recognition task.
Despite the remarkable contributions in theory, the recurrent attention models still suffer from several drawbacks in practice.
First, they only result in small performance improvement.
For instance, Sermanet \etal~\cite{sermanet2014attention} further extend~\cite{ba2014multiple} to fine-grained recognition but only achieve 76.8\% mean accuracy percentage (with 3 glimpses) on Stanford Dogs dataset while the result of GoogLeNet baseline~\cite{szegedy2015going} is 75.5\%.
Second, the computational burden is high.
Calculating features at each glimpse in~\cite{sermanet2014attention} requires forwarding GoogLeNet three times, leading to very slow training and testing.
An exceptional work is Spatial Transformer Networks~\cite{jaderberg2015spatial}, which build on a differentiable attention mechanism that does not need reinforcement learning for training.
As an alternative approach, we show reinforcement learning can still be effective and efficient in improving fine-grained recognition.






\section{Fully Convolutional Attention Networks}


\begin{figure*}[t]
\begin{center}
\includegraphics[width=\linewidth]{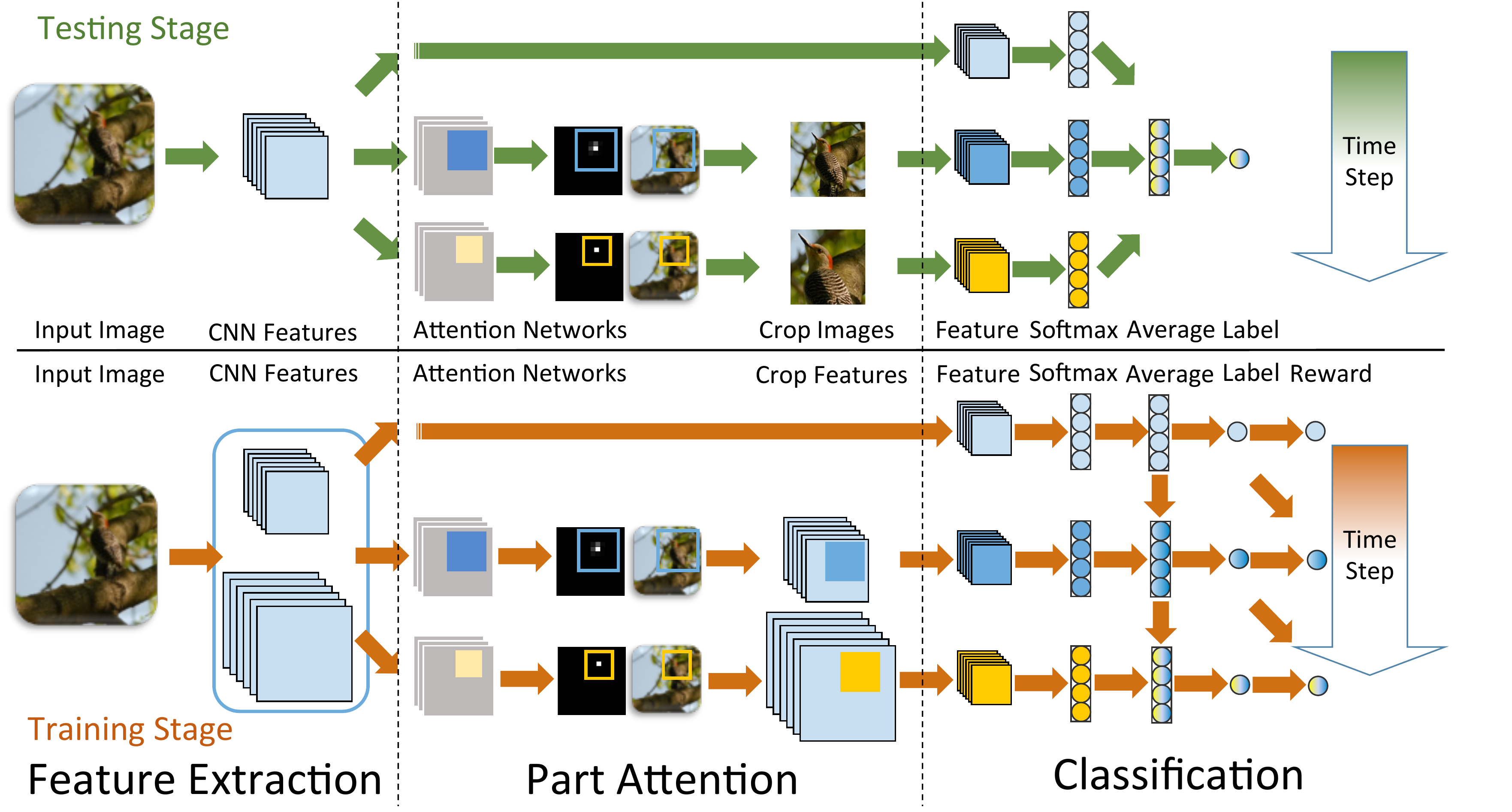}
\end{center}
\caption{The architecture of our FCANs framework.
In this example, the attention network finds two parts of different sizes (the blue region and the yellow region).
The upper part shows the architecture for testing, and the lower part shows the architecture for training.
During testing, we crop all corresponding part patches from the high resolution image for classification.
During training, we re-use the convolutional features in the attention networks for classification.
Note that during testing, we can compute all part attentions simultaneously, which makes the model computationally more efficient than traditional recurrent attention models.
}
\label{fig:architecture}
\vspace{-8pt}
\end{figure*}

Fig.~\ref{fig:architecture} illustrates the architecture of the Fully Convolutional Attention Networks (FCANs) with three main components: the feature network, the attention network, and the classification network.

\textbf{Feature Map Extraction:}
The feature network contains a fully convolutional network that extracts features from the input image and its subsequent attention crops.
These feature maps are shared for both part attention and fine-grained classification.
During experiment, we adopt one of the popular CNN architectures (e.g., VGG-16~\cite{simonyan2014very}, GoogLeNet~\cite{szegedy2015going} or ResNet~\cite{he2016deep}) as the basis fully convolutional network, pre-trained on ImageNet dataset~\cite{wang2014object} and fine-tuned on the target fine-grained dataset.
During testing, the image and all attention crops are resized to a canonical size before feature extraction, similar to~\cite{mnih2014recurrent}.
Hence the amount of computation it performs can be controlled independently of the input image size.

During training, although cropping local image regions can achieve good performance, it requires us to perform multiple forward and backward passes of a deep convolutional network in one batch, where the time complexity for feature extraction depends on the number of parts and number of attention regions sampled for each part.
In practice, this is too time-consuming.
Thus, we extract feature maps from the original image at multiple scales and re-use them across all time steps.
The features for each part is obtained by selecting the corresponding region in the convolutional feature maps, so that the receptive field of the selected region is the same as the size of the part.
As a result, we only need to run the forward pass once in one training batch.

\textbf{Fully Convolutional Part Attention:}
The attention network localizes multiple parts by generating multiple part score maps from the basis convolutional feature maps.
Each score map is generated using two stacked convolutional layers and one spatial softmax layer.
The first convolutional layer uses 64 $3\times3$ kernels, and the second one uses one $3\times3$ kernels to output a single-channel confidence map.
The spatial softmax layer converts the confidence map into probability.
During testing, the model selects the attention region with the highest probability as the part location.
During training, the model samples attention regions multiple times according to the probability map.
The same process is applied for a fixed number of time steps for multiple part locations.
Each time step generates the location for a particular part.
We will detail this step in the following sections.

\textbf{Fine-Grained Classification:}
The classification network contains a convolutional network for each part as well as the whole image.
The classification network for each part is a fully convolutional layer followed by a softmax layer.
Different parts might have different sizes, and a local image region is cropped around each part location according to its size.
The final prediction score is the average of all the prediction scores from the individual classifiers.
In order to discriminate the subtle visual differences, each local image region is cropped at high resolution.




\subsection{Model}
The entire attention problem is formulated into a Markov Decision Process (MDP).
During each time step of MDP, the FCANs work as an agent to perform an action based on the observation and receives a reward.
In our work, the action corresponds to the location of the attention region, the observation is the input image and the crops of the attention regions and the reward measures the quality of the classification using the attention region.
The target of our learning is to find the optimal decision policy to generate actions from observations, characterized by the parameters of the FCANs, to maximize the sum expected reward across all time steps.

We define the input image as $x$ and the feature network (parameterized by $\theta_f$) computes the feature maps as $\phi(x, \theta_f)$.
The attention network outputs $T$ attention locations $\{l^1, \ldots, l^T\}$ with each location $l^t \sim \pi(l^t | \phi, \theta^t_l)$, where $\pi$ is the policy for attention selection, parameterized by $\theta_l=\{\theta^t_l\}_{t=1\cdots T}$.
At time step $t$, the classification component crops an image region at location $l^t$, extracts a new feature $\phi(l^t)$ and predicts classification score $s_t$ with the classification network (parameterized by $\theta_c=\{\theta^t_c\}_{t=1\cdots T}$).
It then computes the final classification score $S_t$ as the average of all prediction scores until time $t$
\begin{equation}
S_t = \frac{1}{t} \sum_{\tau=1}^t s_{\tau}(\phi(l^{\tau}), \theta^{\tau}_c)
\end{equation}
Note that in FCANs, both $\theta_l$ and $\theta_c$ have different sets of parameters $\{\theta^t_l, \theta^t_c\}$ at different time steps.
Only the parameters in the feature network $\theta_f$ are shared across all time steps.
This is different from the original recurrent attention models~\cite{mnih2014recurrent, ba2014multiple} where all parameters are shared across multiple time steps.
The reward $r^t$ for the $t$-th step measures how the output of $S_t$ matches the ground truth label $y$.






\subsection{Training}

Since there are no ground-truth annotations to indicate where to select attention regions and each attention is a non-differentiable function, we adopt reinforcement learning to learn the network parameters.

Given a set of training images with ground truth labels $(x_n, y_n)_{n=1\cdots N}$, we jointly optimize the three components to maximize the following objective function:
\begin{equation}
\max_{\theta} J(\theta) =  \max_{\theta_f, \theta_l, \theta_c} R(\theta_f, \theta_l) - L(\theta_f, \theta_c)
\end{equation}
where $\theta = \{\theta_f, \theta_l, \theta_c\}$ are the parameters of the feature networks, the attention networks and the classification networks respectively.
\begin{equation}
L(\theta_f, \theta_c) = \frac{1}{NT} \sum_{n=1}^N \sum_{t=1}^T L^t_n(x_n, y_n, \theta_f, \theta_c)
\end{equation}
is the average cross-entropy classification loss over $N$ training samples and $T$ time steps.
\begin{equation}
R(\theta_f, \theta_l) = \frac{1}{NT} \sum_{n=1}^N \sum_{t=1}^T \mathbb{E}_{\theta}[r^t_n]
\end{equation}
is the average expected reward over $N$ training samples and $T$ time steps.
\begin{equation}
\mathbb{E}_{\theta}[r^t_n] = \sum_{l^t_n} \pi(l^t_n|x_n, \theta_f, \theta^t_l) \ r^t_n
\end{equation}
is the expected reward of the $t$-th selected attention region from the $n$-th sample.
$\theta^t_l$ is the parameters of the $t$-th attention network,
$\pi(l^t_n|x_n, \theta_f, \theta^t_l) = \pi(l^t_n|\phi(x_n), \theta^t_l)$ is the probability of selecting $l^t_n$ as the attention region.
The reward function $r^t_n$ is crucial for developing an efficient learning algorithm. We describe the design of the reward function in the following section.

\subsection{Reward Strategy} \label{sec: reward}

A straightforward reward strategy is to measure the quality of the attention region selection policy as a whole using the final classification result, i.e., $r^t_n = 1$ if $t=T$ and $y_n=\arg\max_y P(y | S^T_n)$, and 0 otherwise.
Although MDP with such a reward strategy can learn in a recurrent way~\cite{mnih2014recurrent}, it confuses the effect of the selected regions in different time steps,
and it might lead to the problem of convergence difficulty.

We consider an alternative reward strategy, namely {\em greedy reward}:
\begin{equation}
r^t_n=\left\{
\begin{array}{cc}
1 & t = 1 \wedge y_n = \arg\max_y P(y | S^1_n) \\
1 & t > 1 \wedge y_{n} = \arg\max_y P(y | S^t_n) \wedge L^t_n < L^{t-1}_n \\
0 & otherwise \end{array}
\right.
\end{equation}
where $L^t_n$ is the classification loss for the $n$-th sample at $t$-th step.
If the image is classified correctly in the first step, the attention network immediately receives a reward.
In other time steps, we reward the corresponding attention network only if the image is classified correctly and the classification loss decreases with regards to the last time step.
Otherwise, the attention network receives zero reward.
Since the attention network immediately receives a reward when an image is correctly classified with the current attention region, the convergence of training is much easier.

\subsection{Optimization}

\begin{figure}[t]
\begin{center}
\subfigure[Forwarding]{
\includegraphics[scale = 0.42]{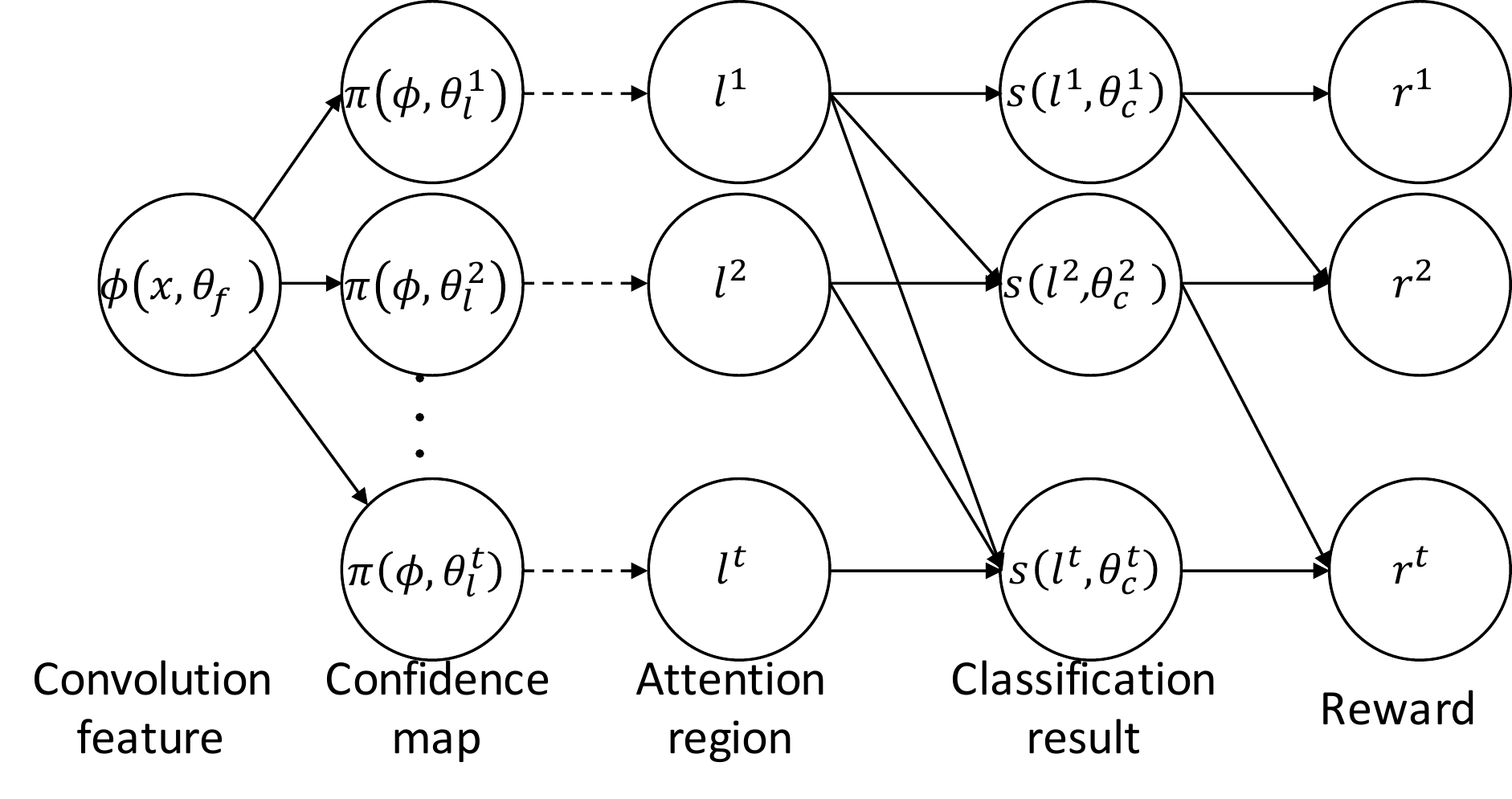}}
\hspace{0.05in}
\subfigure[Back-propagation]{
\includegraphics[scale = 0.42]{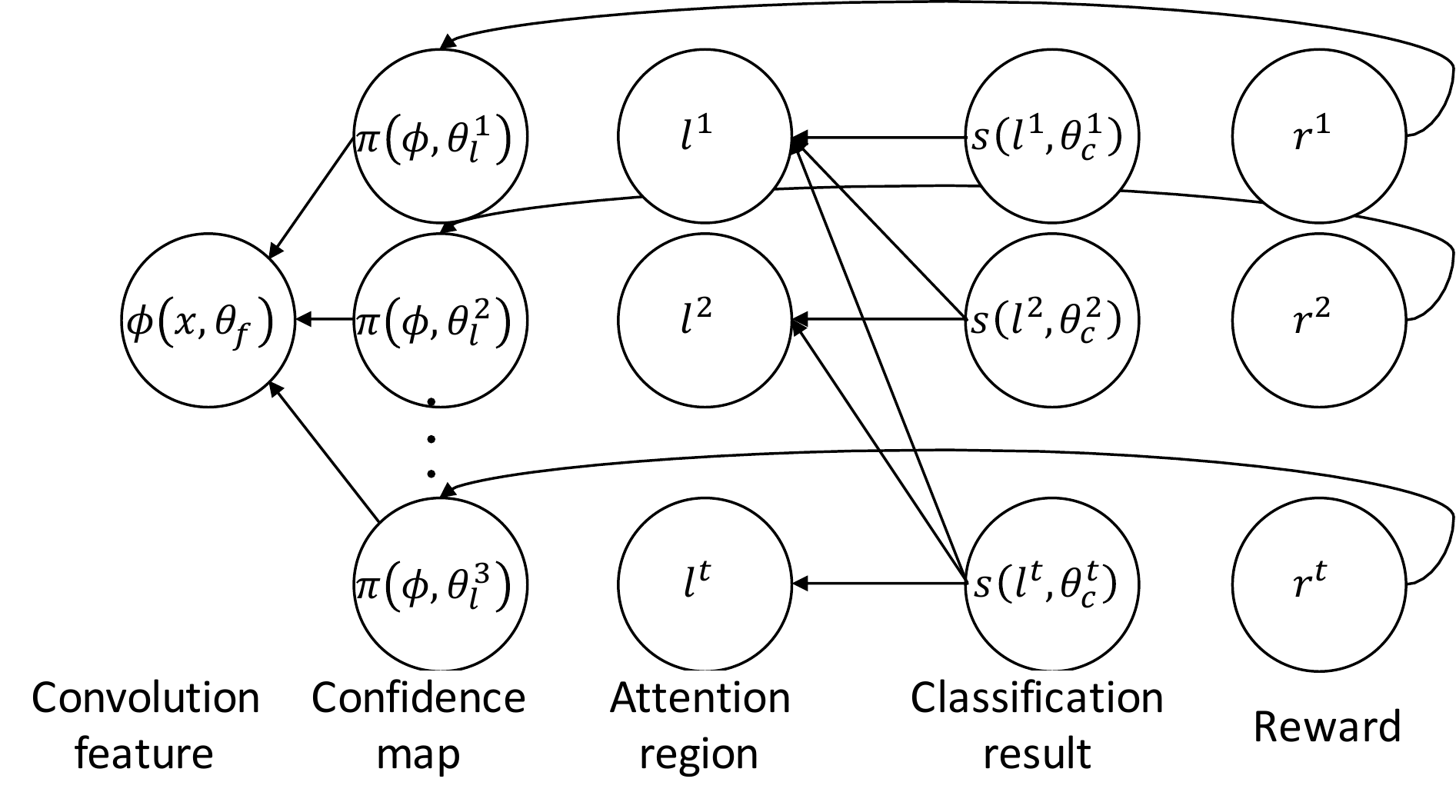}}
\end{center}
\caption{The forward (a) and back-propagation (b) processes for training attention networks as MDPs.
The dashed lines indicate the sampling procedures.
}
\label{fig:optimization}
\end{figure}

It is difficult to directly compute the gradient of $\mathbb{E}_{\theta}[r^t_n]$ over $\theta$ because it requires evaluating exponentially many possible part locations during training.
Hence we employ REINFORCE algorithm and approximate the gradient in a Monte Carlo way~\cite{sutton1999policy}.

\begin{equation}
\nabla_{\theta} \mathbb{E}_{\theta}[r^t_n] \approx \frac{1}{K} \sum_{k=1}^K \nabla_{\theta} \left(\log \pi(l^t_{nk} | \phi(x_n), \theta^t_l)\right) r^t_{nk}
\label{eq:policy_gradient}
\end{equation}
where $l^t_{nk}\sim \pi(l^t_n | \phi(x_n), \theta^t_l) $ is sampled according to a multinomial distribution parameterized by the output confidence map of the $t$-th attention network.

The forward process of training the attention networks as MDPs is shown in Fig.~\ref{fig:optimization}. Given the basis convolutional feature maps $\phi(x)$ as input, the attention networks output the confidence map $\pi(\phi, \theta^t_l)$ at different time step $t$. Each $\pi(\phi, \theta^t_l)$ forms a multinomial distribution, and the location of attention region $l^t$ is sampled under the distribution. The sampling procedure is repeated for $K$ times. We then use them for classification network and further get the reward $r^t$.

During back-propagation, the gradient $\nabla_{\theta} L(\theta_f, \theta_c)$ can be  obtained by back-propagating the classification networks.
The gradient $\nabla_{\theta} R(\theta_f, \theta_l)$ is calculated using policy gradient as shown in Equation~\ref{eq:policy_gradient}.
Notice that when the reward is 0, we can just ignore the sample.

\subsection{Implementation Details}

\textbf{Step-wise training:} Although jointly training the entire model is possible, we develop a 3-step algorithm for the sake of training speed.
In the first step, we initialize and fine-tune the CNN model to extract the basis convolutional feature maps for attention and classification.
In the second step, we fix and cache the basis convolutional feature maps from the first step, and train the attention networks separately.
In the third step, we fix and cache the selected attention regions from the second step, and fine-tune the final classification model.
Through feature caching, repeated feature calculating is avoided.
Notice that the convolutional neural networks for attention and the final classification is different, though they are initialized similarly during pre-training described below.
We repeat these steps several times until convergence.

\textbf{Fast-RCNN approximation:} During training, although we can compute a multi-scale feature maps to obtain the features for high resolution region crops. It could still be time-consuming when the image resolution is large.
Thus, we employ an approximated feature extraction method that is similar to Fast-RCNN~\cite{girshick2015fast}, where we only compute a feature map from the input image at one scale.
The convolutional features for each part is obtained by selecting the corresponding region in the convolutional feature map of the whole image, so that the receptive field of the selected region is the same as the size of the part.
This further accelerates the training of attention networks.

Note that since we adopt a 3-step training, the Fast-RCNN approximation are only utilized during attention network training.
The final classification networks are still trained given the features extracted from cropped high resolution images.

\subsection{Discussion}

Our attention component is inspired from the recurrent visual attention model~\cite{mnih2014recurrent}.
However, instead of building a recurrent attention network that share parameters over different time steps, our model uses multiple convolutional networks with different parameters to model the temporal effect.
During testing, these attention networks work like independent part detectors that share the same basis image feature.
It is even possible to combine all the attention networks into a single convolutional network to compute part attentions simultaneously.
This makes inference much faster.




\section{Experiments}

\begin{table}[t]
\begin{center}
\rowcolors{2}{}{yelloworange!25}
\addtolength{\tabcolsep}{2.5pt}
\begin{tabular}{l c c c c c}
\toprule[0.2 em]
Dataset & \#Class & \#Train & \#Test & BBox & Part \\
\toprule[0.2 em]
Stanford Dogs~\cite{khosla2011novel} \ & 120 & 12,000  &  8,580 & $\surd$ &    \\
Stanford Cars~\cite{krause20133d} \ & 196 & 8,144  & 8,041 & $\surd$ &  \\
CUB-200-2011~\cite{wah2011caltech} \ & 200 & 5,994 & 5,794 & $\surd$ & $\surd$ \\
Food-101~\cite{bossard2014food} \ & 101 & 75,750 & 25,250 & $ $ & $ $ \\
\bottomrule[0.1 em]
\end{tabular}
\vspace{1pt}
\caption{Statistics for the four fine-grained benchmark datasets.}
\label{tab:statistics}
\end{center}
\end{table}

We conduct extensive experiments on four benchmark datasets, including CUB-200-2011~\cite{wah2011caltech}, Stanford Dogs~\cite{khosla2011novel}, Stanford Cars~\cite{krause20133d}  and Food-101~\cite{bossard2014food}.
Table~\ref{tab:statistics} shows the statistics of the four datasets.

\subsection{Experimental Setup}
We use the ResNet-50~\cite{he2016deep} for feature extraction. During pre-training, we first resize all images to $512 \times512 $ resolution, and fine-tune the ResNet-50 with randomly cropped $448\times448$ patches.
For each input image, ResNet-50 outputs a $2048\times16\times16$ \texttt{res\_5c} feature map.
We then use the feature map to train the attention networks to find two parts.
The first part selects a $4\times4$ region in the feature map (corresponding to a $128\times128$ patch in the resized image), and the second one selects a $8\times8$ region (corresponding to a $256\times256$ patch in the resized image).
We then crop the two result attention patches and resize to $512\times512$ to train ResNet-50 prediction models in the final classification stage.

All models are trained using RMSProp with batch size of 512 and 90 epochs.
The initial learning rate is 0.01 and multiplied by 0.1 every 30 epochs.
Our implementation is based on Caffe~\cite{jia2014caffe}.

\subsection{Computational Time}

On Stanford Dogs dataset, our FCANs take 3 hours to train on a single Tesla K40 GPU, significantly faster than a conventional recurrent attention model~\cite{sermanet2014attention} that takes about 30 hours to converge in our implementation.
Fine-tuning the convolutional features requiring additional training time for both models.
For an image with 512$\times$512 resolution, our testing time is $\sim$150ms.
The cost of attention selection is negligible compared with the feature calculation time.
Compared with recurrent attention models~\cite{sermanet2014attention} that takes $\sim$250ms during testing, our method is faster.

\subsection{Comparison with State-of-the-Arts}

We compare our framework with all previous methods and summarize the results from Table~\ref{tab:dog} to Table~\ref{tab:food}.

On CUB-200-2011, our recognition accuracy (84.3\%) is comparable with all state-of-the-art methods~\cite{lin2015bilinear, jaderberg2015spatial, kong2016low} without using ground-truth bounding boxes during testing.



On Stanford Dogs, Stanford Cars and Food-101, our model is also very competitive.
For example, we obtain 93.1\% accuracy on Stanford Cars test set with bounding box during testing, which is so far the best result.
Note that our baseline method Sermanet \etal~\cite{sermanet2014attention} uses reinforcement learning based recurrent attention models, which is similar to our approach.
Our method improves them by more than 12\% on Stanford Dogs, suggesting the FCANs as an effective framework for fine-grained recognition.

\begin{table}[t]
  \centering
  \rowcolors{2}{}{yelloworange!25}
  \addtolength{\tabcolsep}{2.5pt}
    \begin{tabular}{l c c}
      \toprule[0.2 em]
      {\bf CUB-200-2011} & Accuracy(\%) & Acc w. Box(\%) \\
      \toprule[0.2 em]
      \midrule
      Zhang \etal~\cite{zhang2014part} & 73.9 & 76.4 \\
      Branson \etal~\cite{branson2014bird} & 75.7 & 85.4$^*$ \\
      Simon \etal~\cite{simon2015neural} & 81.0 & - \\
      Krause \etal~\cite{krause2015fine} & 82.0 & 82.8 \\
      Lin \etal~\cite{lin2015bilinear} & 84.1 & 85.1 \\
      Jaderberg \etal~\cite{jaderberg2015spatial} & 84.1 & - \\
      Kong \etal~\cite{kong2016low} & 84.2 & - \\
      \midrule
      Our Model & {\bf 84.3} & 84.7 \\
      \bottomrule[0.1 em]
    \end{tabular}
    \vspace{1pt}
    \caption{Comparison to related work on CUB-200-2011 dataset. $^*$ Testing with both ground truth box and parts.}
    \label{tab:bird}
\end{table}

\begin{table}
  \centering
  \rowcolors{2}{}{yelloworange!25}
  \addtolength{\tabcolsep}{2.5pt}
    \begin{tabular}{l c c}
      \toprule[0.2 em]
      {\bf Stanford Dogs} & Accuracy(\%) & Acc w. Box(\%) \\
      \toprule[0.2 em]
      \midrule
      Gavves \etal~\cite{gavves2013fine} & - & 50.1 \\
      Simon \& Rodner~\cite{simon2015neural} & 68.1 & - \\
      Sermanet \etal~\cite{sermanet2014attention} & 76.8 & - \\
      Zhang \etal~\cite{zhang2015weakly}  & 79.9 & - \\
      Krause \etal~\cite{krause2016unreasonable} & 82.6 & - \\
      \midrule
      Our Model & {\bf 88.9} & - \\
      \bottomrule[0.1 em]
    \end{tabular}
    \vspace{1pt}
    \caption{Comparison to related work on Stanford Dogs dataset.}
    \label{tab:dog}
\end{table}

\begin{table}
  \centering
  \rowcolors{2}{}{yelloworange!25}
  \addtolength{\tabcolsep}{2.5pt}
    \begin{tabular}{l c c}
      \toprule[0.2 em]
      {\bf Stanford Cars} & Accuracy(\%) & Acc w. Box(\%) \\
      \toprule[0.2 em]
      \midrule
      Chai et al.~\cite{chai2013symbiotic} & 78.0 & - \\
      Gosselin et al.~\cite{gosselin2014revisiting} & 82.7 & 87.9 \\
      Girshick et al.~\cite{girshick2014rich} & 88.4 & - \\
      Lin et al.~\cite{lin2015bilinear} & 91.3 & - \\
      Wang et al.~\cite{wang2016mining} & - & 92.5 \\
      Krause et al.~\cite{krause2015fine} & 92.6 & 92.8 \\
      \midrule
      Our Model & 91.5 & {\bf 93.1} \\
      \bottomrule[0.1 em]
    \end{tabular}
    \vspace{1pt}
    \caption{Comparison to related work on Stanford Cars dataset.}
    \label{tab:car}
\end{table}

\begin{table}[t]
\centering
\rowcolors{2}{}{yelloworange!25}
\addtolength{\tabcolsep}{2.5pt}
\begin{tabular}{c c c}
\toprule[0.2 em]
Method & Accuracy(\%) & Acc w. Box(\%) \\
\toprule[0.2 em]
L. Bossard \etal ~\cite{bossard2014food} & 50.8 & - \\
A. Myers \etal  ~\cite{meyers2015im2calories} & 79.0 & - \\
\midrule
Our Model & {\bf 86.3} & - \\
\bottomrule[0.1 em]
\end{tabular}
\vspace{1pt}
\caption{Experimental results on Food-101 dataset.}
\label{tab:food}
\end{table}

\subsection{Ablation Study}

\textbf{Effect of Attention:}
Since our approach is roughly three times (full image + two attention regions) more expensive than a single model during testing, we conduct two additional model-fusion baselines to demonstrate its superiority.
One is the random region experiment, where we augment the baseline single image model with two random cropped regions.
The second baseline is the center region experiment, where we crop two center regions in the image.
The sizes of the two crops in both experiments are the same as the sizes of the parts in the attention model.
Table~\ref{tab:effect_of_attention} summarizes the results.
When costing the same amount of testing time, the attention networks clearly outperform random region and center region models.

\begin{table}[t]
\centering
\rowcolors{2}{}{yelloworange!25}
\addtolength{\tabcolsep}{2.5pt}
\begin{tabular}{c c c c c}
\toprule[0.2 em]
Method & \ Dogs \ & \ Cars \ & \ Birds \ & \ Foods \\
\toprule[0.2 em]
Finetune baseline \ & 87.3 & 89.7 & 82.0 & 82.1\\
+ Random regions \ & 87.9 & 90.1 & 82.3 & 83.0 \\
+ Center regions \ & 87.5 & 90.6 & 82.4 & 82.7 \\
+ Attention regions \ & \bf{88.9} & \bf{91.5} & \bf{84.3} & \bf{86.3} \\
\bottomrule[0.1 em]
\end{tabular}
\vspace{1pt}
\caption{Experimental comparison on the effect of attentions.}
\label{tab:effect_of_attention}
\end{table}

\textbf{Number of Attentions:}
Table~\ref{tab:number_of_attention} summarizes the results of how the number of attentions affects the final classification accuracy.
Take Stanford Dogs as an example, after fine-tuning the baseline ResNet-50 achieves 87.3\% accuracy.
Combining one $8\times8$ attention region with the prediction results of original image improves significantly to 88.5\%.
Combining one $8\times8$ region, one $4\times4$ region and the original image together further improves the results to 88.9\%.
We find adding more than two attentions (i.e. 3 attentions) only improves the performance slightly at the expense of more computations.
Hence throughout the experiments we fix the number of attentions as two.

\begin{table}[t]
\centering
\rowcolors{2}{}{yelloworange!25}
\addtolength{\tabcolsep}{2.5pt}
\begin{tabular}{c c c c c}
\toprule[0.2 em]
Method & \ Dogs \ & \ Cars \ & \ Birds \ & \ Foods \\
\toprule[0.2 em]
Finetune baseline \ & 87.3 & 87.5 & 82.0 & 82.1\\
One attention only \ & 88.1& 84.2 & 80.4 & 79.9 \\
+ One attention & 88.5 & 90.2 & 83.3 & 85.5 \\
+ Two attentions  \ & 88.9 & 91.5 & {\bf84.3} & 86.3 \\
+ More attentions \ & {\bf 89.0} & {\bf 91.6} & {\bf 84.3} & {\bf 86.5} \\
\bottomrule[0.1 em]
\end{tabular}
\vspace{1pt}
\caption{Experimental comparison on the number of attentions.}
\label{tab:number_of_attention}
\end{table}

\textbf{Reward Strategy:}
Table~\ref{tab:reward_strategy} illustrates the effectiveness of our training reward strategy.
Compared against the traditional reward setting which only assigns a reward after all attention iterations, our greedy reward strategy works significantly better.
We hypothesize that the greedy reward helps the reinforcement learning to quickly converge to discriminative sub-regions.

\begin{table}[t]
\centering
\rowcolors{2}{}{yelloworange!25}
\addtolength{\tabcolsep}{2.5pt}
\begin{tabular}{c c c c c}
\toprule[0.2 em]
Method & \ Dogs \ & \ Cars \ & \ Birds \  & \ Foods \\
\toprule[0.2 em]
Baseline reward \  & 88.1 &  90.5  &  82.9  &   84.7 \\
Greedy reward \ & \bf{88.9} & \bf{91.5} & \bf{84.3} & \bf{86.3} \\
\bottomrule[0.1 em]
\end{tabular}
\vspace{1pt}
\caption{Experimental comparison on the reward strategy.
The baseline reward strategy only assigns a reward after all attention iterations.}
\label{tab:reward_strategy}
\end{table}

\begin{figure*}[t]
\begin{center}
\includegraphics[scale=0.8]{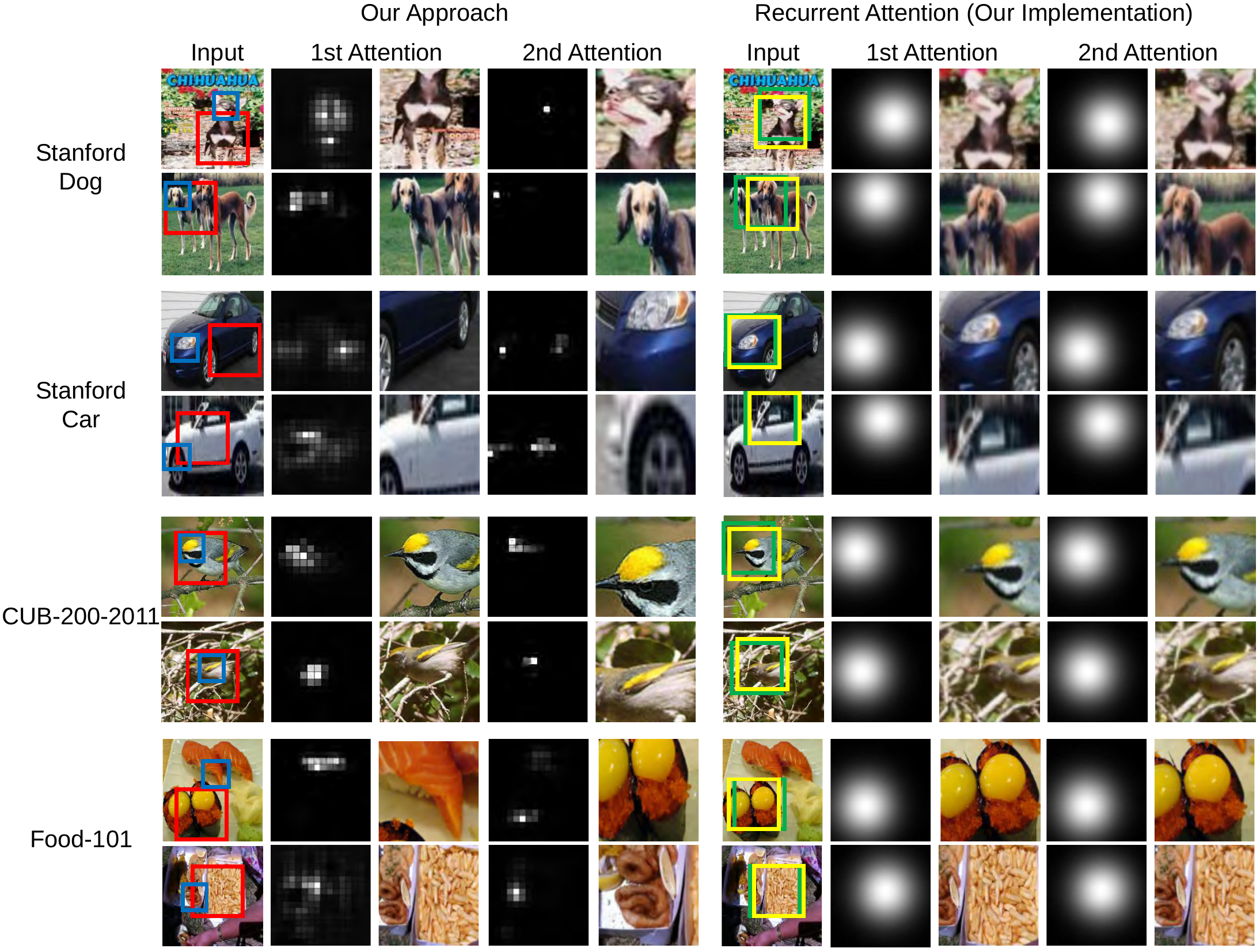}
\end{center}
\caption{Qualitative comparison between our method (left) and recurrent attention~\cite{sermanet2014attention} (right) on different datasets.
On the left, we plot the first two attention regions regenerated by FCAN, which corresponds to $4\times4$ and $8\times8$ attention regions respectively (lighter color indicates higher score).
On the right, we also show the first two selected regions by~\cite{sermanet2014attention} using our implementation.
}
\label{fig:attention_illustration}
\vspace{-8pt}
\end{figure*}

\subsection{Qualitative Results}


We qualitatively compare the attention regions selected by our model and the recurrent attention model~\cite{sermanet2014attention} in Fig.~\ref{fig:attention_illustration}.
Both models contain attention mechanisms and apply reinforcement learning to train to focus on local discriminative regions.
We observe that in our model different attentions correspond to different image regions, while the attention regions generated in~\cite{sermanet2014attention} focus on only one region.
Our attention map is also more diverse than the attention map in~\cite{sermanet2014attention}.
This illustrates how our attention model outperforms the previous reinforcement learning based attention work.



\section{Conclusion}
In this paper, we present Fully Convolutional Attention Networks (FCANs) for fine-grained recognition.
With the fully convolutional architecture, our model is much faster than previous reinforcement learning based visual attention models during both training and testing.
We conduct extensive experiments on four different fine-grained benchmark datasets and show its competitive performance against state-of-the-art methods.

\small
\bibliographystyle{./IEEEtran}
\bibliography{./egbib}
\end{document}